\documentclass[letterpaper, 10 pt, conference]{ieeeconf}  

\usepackage{graphicx}
\usepackage{flushend}
\usepackage{balance}

\IEEEoverridecommandlockouts                              

\title{\LARGE \bf
Towards an Efficient Synthetic Image Data Pipeline for Training Vision-Based Robot Systems
}

\author{Peter Gavriel$^{1}$, Adam Norton$^{1}$, Kenneth Kimble$^{2}$, and Megan Zimmerman$^{2}$
\thanks{*This work was supported by the National Institute of Standards and Technology (NIST) under awards 70NANB22H114 and 70NANB24H047.}
\thanks{$^{1}$New England Robotics Validation and Experimentation (NERVE) Center, University of Massachusetts Lowell, Lowell, MA, USA
        {\tt\small peter\_gavriel,adam\_norton@uml.edu}}%
\thanks{$^{2}$National Institute of Standards and Technology (NIST), Gaithersburg, MD, USA
        {\tt\small kenneth.kimble,megan.zimmerman@nist.gov}}%
}

\begin{document}

\maketitle
\thispagestyle{empty}
\pagestyle{empty}




\section{Introduction and Background}

The sophistication and accuracy of computer vision systems, such as those used for object detection, pose detection, and scene segmentation, directly enables the proficiency of robotics applications across many domains including grasping, assembly, and navigation. 
Robotic manipulation of objects in an unstructured environment or during failure recovery often requires an understanding of the objects in the workspace, such as the object's pose, which relies on its computer vision capabilities. 
Computer vision systems are primarily built using neural network architectures that rely on quality training data in order to achieve high levels of performance. Many large data sets of labeled images have been assembled over the years to train and benchmark computer vision tasks. However, these manually collected and labelled datasets have proven to be resource intensive to create, leave room for human error in labeling, and the data diversity necessary for robust performance on complex vision tasks is increasingly difficult to achieve.

As an alternative, neural networks can be trained on synthetic data generated from simulations, which has major advantages over real-world data, including much faster generation speed, perfect ground truth labels, and limitless options for automating the addition of variability (e.g., noise, lighting, camera angles). In many cases, models trained on synthetic data have failed to reach adequate performance when tested with real-world data (e.g., sim-to-real transfer applications). However, due to advances in the tools used to generate synthetic data, as well as developed techniques such as domain randomization~\cite{Tremblay2018DR} and domain adaptation~\cite{Wang2018DASurvey}, the performance gap between real-world and synthetic data continues to narrow. Research has shown it to be possible to achieve state of the art performance utilizing domain randomized synthetic data for industrial object detection tasks~\cite{heindl2020blendtorch}, as well as 6D pose estimation~\cite{zakharov2022photo,hagelskjaer2022parapose}. Assuming this trend continues, synthetic training data may soon become prevalent enough to warrant infrastructure dedicated to generating high quality synthetic datasets quickly and easily, which enables the shortening of development cycles for computer vision applications.  

To this end, we propose a framework for a synthetic image data generation pipeline that includes real-world data capture of target objects, digital reconstruction of the objects into 3D models, and generates labeled synthetic data primed for learning-based robotics applications. 
While solutions for each of these components exist in some form already, we propose an efficient end-to-end pipeline that combines them as interchangeable parts that can take different forms and evolve as new software tools become available.

\section{Pipeline Framework}


\subsection{Consistent Real-World Data Capture}

Having consistent image data with known ground truth poses is an ideal output to lead into digital reconstruction. It is possible to compute the camera poses for sequences of unlabelled images using Perspective-n-Point (PnP) algorithms, but they can be inconsistent and require the tuning of parameters to find best results which can be time consuming. We leverage an apparatus developed as part of the NIST Manufacturing Objects and Assemblies Dataset (MOAD) project~\cite{nistmoadweb} to collect data in an automated fashion utilizing a motorized turntable surrounded by 5 sensor modules each containing a DSLR camera and an Intel RealSense depth camera at specified heights and angles; an evolution of the BigBIRD data collection system~\cite{singh2014bigbird}. The form of data captured from the sensor modules and the angular resolution of the turntable can be configured as needed. The data collection process is completely hands off once started, and a full 360 degree high resolution image scan of an object at 5 degree increments can be completed in 5 minutes. The output of this apparatus can provide RGB images, depth images, and point clouds all with known transformations with respect to the center of the turntable which can immediately be used for digital reconstruction. 
To date, NIST-MOAD consists of the NIST Assembly Task Boards (ATB) 1 through 4~\cite{kimble2022performance}, which are robot manipulation and perception benchmarking tools that include small parts assembly and disassembly tasks, with RGB images and point clouds of the individual components, the assembled boards, and the disassembled boards~\cite{nistmoadweb}.

Carefully characterizing this component of the pipeline is important for the sake of reproducibility, making fair comparisons between techniques for digital reconstruction, and troubleshooting downstream problems in the synthetic data pipeline.


\subsection{Accurate Digital Reconstruction}

\paragraph*{Photogrammetry}Structure from motion (SFM) photogrammetry has been a popular solution for digital reconstruction, especially for applications like mapping with unmanned aerial systems (UAS), geological reconstruction, and construction site management. To achieve desirable results with SFM reconstruction, the initial feature extraction and feature matching is a critical component. It can be difficult to extract a sufficient number of features from objects with minimal texturing or objects with simple geometry, and SFM can struggle to properly match features for objects with symmetries~\cite{onur2017sfm}. Bianco et. al~\cite{bianco2018sfm} provide a good overview of the different configurations of the SFM pipeline and how performance is effected. 

\paragraph*{Structured Light 3D Scanning}All-in-one handheld 3D scanning systems are also commercially available and provide a combined off the shelf solution for real-world data capture and digital reconstruction. Devices such as the Einscan-SE have already been utilized to generate 3D models for the Household Objects for Pose Estimation (HOPE) dataset, and the explicit purpose for these models is to enable synthetic data generation~\cite{tyree20226}. While convenient, these kinds of handheld 3D scanners tend to have difficulty tracking objects with reflective or transparent materials, which are extremely common characteristics for objects in medical and industrial applications. 

\paragraph*{Radiance Fields}Neural Radiance Fields (NeRFs) have emerged as a novel method of synthesizing novel viewpoints of complex scenes from a sparse image set with impressive results that are less expensive to produce than traditional SFM photogrammetry, contain less artifacts, and capture small details, textures, and light effects very well~\cite{mildenhall2021nerf}. Rabby et. al~\cite{rabby2023beyondpixels} provide a comprehensive review of the many iterations and improvements made to NeRFs in the past few years. In addition to the many improvements and variations of NeRFs that were introduced, NeRF Studio was created as an open-source modular framework to streamline the process of NeRF development~\cite{tancik2023nerfstudio}. Then in 2023, yet another new method for novel view synthesis; 3D Gaussian Splatting (3D GS), was introduced which had similarly impressive fidelity as NeRFs but with a rendering efficiency increase of several orders of magnitude over traditional NeRFs~\cite{kerbl3Dgaussians}. Since their introduction, papers utilizing and making improvements on 3D GS have begun to dominate the literature due to their advantages over NeRFs such as rendering speed and allowing unprecedented editing and control over scenes~\cite{chen2024survey}. NeRF Studio has since been developed to allow for the creation of 3D GS in addition to NeRFs, likely making it the best starting point for development with either reconstruction method. 

\paragraph*{Model Extraction}The reconstruction must be exported in a way that preserves all of the relevant details so that it may then be loaded into the data generator. This includes being able to properly segment the object of interest from any background data that was reconstructed. Additionally, the exported model should be closed, and should represent all sides of the object. For example, if the bottom of the object is algorithmically filled in rather than constructed using captured data, then synthetic data generated with views of the bottom of the object may cause performance issues as it would not properly represent the object in training. In the case of using NeRFs for reconstruction, work has already been done to fine tune mesh creation from NeRF representations~\cite{yariv2023bakedsdf}, and to create editable UV texture maps from neural representations~\cite{srinivasan2023nuvo}. For 3D GS, several methods have already been published demonstrating high quality 3D mesh extraction~\cite{guedon2023sugar,chen2023neusg}.
A description of the digital reconstruction component should include comprehensive information about the form of the input, the methods and parameters used for reconstruction, and the method and form of the output model.

\begin{figure}[t!]
\centerline{\includegraphics[width=0.45\textwidth]{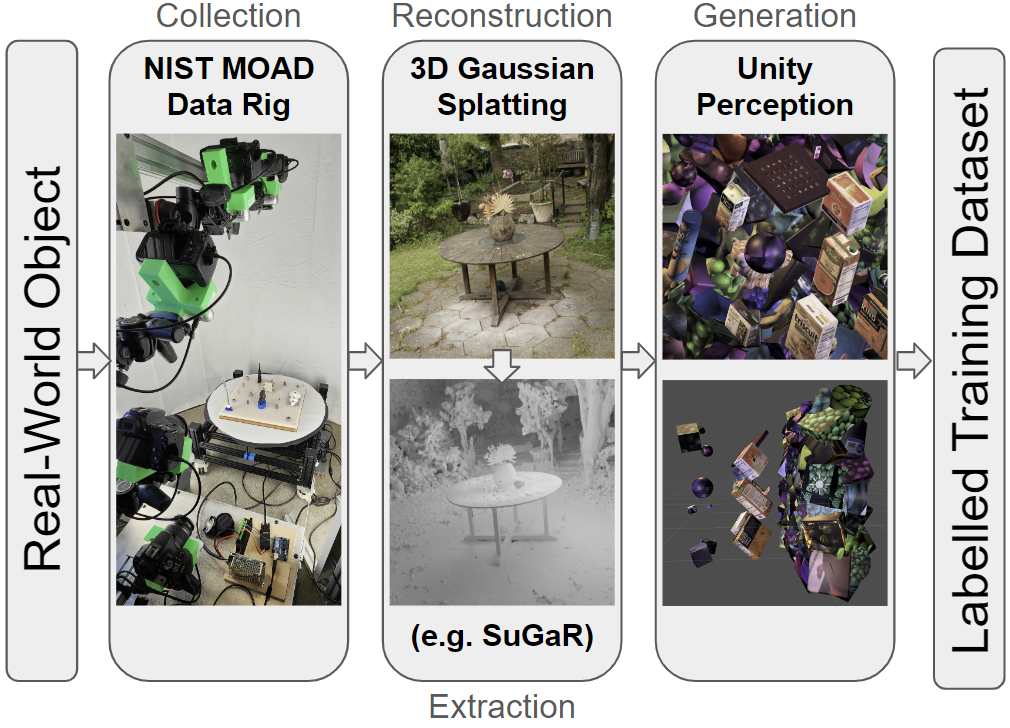}}
\caption{Diagram example configuration of the proposed synthetic image data generation pipeline utilizing the MOAD Data Rig~\cite{kimble2022performance} for data collection, 3D Gaussian Splatting~\cite{kerbl3Dgaussians} and SuGaR~\cite{guedon2023sugar} for digital reconstruction, and Unity Perception~\cite{borkman2021unity} for data generation.}
\label{fig:diagram}
\end{figure}

\subsection{Synthetic Dataset Generation}

Many tools have already been created for scene simulation, and they continue to improve in terms of photo-realism and accurate physics. A review of indoor synthetic data generation in 2024~\cite{Schieber2024} categorizes four approaches to generating synthetic data; crop-out based, graphic API-based, 3D Modeling-based, and 3D game engine-based, with 3D modeling-based and 3D game engine-based appearing to be the most capable and most utilized by researchers in recent years. The most capable, flexible, and extensible tools already available include BlenderProc (3D modeling-based)~\cite{denninger2019blenderproc}, Kubric (3D modeling-based)~\cite{greff2022kubric}, and Unity Perception (3D game engine-based)~\cite{borkman2021unity}, all of which include the capability to incorporate domain randomization for backgrounds, object poses, textures, materials, camera poses, lighting, and distractor objects. Domain randomization in these aspects has already been shown to be important for bridging the sim-to-real performance gap~\cite{Tremblay2018DR}. 
Other research has leveraged NeRFs directly to generate synthetic training data and achieved similar performance to synthetic training data created by BlenderProc~\cite{ge2022neural}.
When being described, the data generator should list all of the parameters being varied in the generation process alongside the range in which they are varied and the type of distribution used unless it is uniformly random. Additionally, the description should include information about the output such as image resolution, and the format of the ground truths being generated.



\bibliographystyle{ieeetr}
\balance
\bibliography{main.bib}

\end{document}